\documentclass[a4paper, 10pt, conference]{ieeeconf}      

\IEEEoverridecommandlockouts                              

\usepackage{graphics} 
\usepackage{epsfig} 
\usepackage{mathptmx} 
\usepackage{times} 
\usepackage{amsmath} 
\usepackage{amssymb}  
\usepackage{xcolor}
\usepackage{multirow}
\usepackage{subcaption}
\usepackage[font={footnotesize,it}]{caption}
\usepackage{cite}
\usepackage{cleveref}
\usepackage{algorithm,algorithmic}
\usepackage{fancyhdr}

\title{\LARGE \bf
Anomaly Detection in Connected and Automated Vehicles using an Augmented State Formulation
}

\author{Yiyang Wang$^{1}$, Neda Masoud$^{1}$, and Anahita Khojandi$^{2}$
\thanks{Accepted to be published in: 2020 Forum on Integrated and Sustainable Transportation Systems (FISTS), Delft, Netherlands, 2020, pp. 156-161.}
\thanks{\copyright 2020 IEEE. Personal use of this material is permitted. Permission from IEEE must be obtained for all other uses, in any current or future media, including reprinting/republishing this material for advertising or promotional purposes, creating new collective works, for resale or redistribution to servers or lists, or reuse of any copyrighted component of this work in other works. doi: 10.1109/FISTS46898.2020.9264885.}
\thanks{$^{1}$Department of Civil Engineering, University of Michigan, Ann Arbor 48109, USA
        {\tt\small yiyangw@umich.edu, nmasoud@umich.edu}}%
\thanks{$^{2}$Department of Industrial and Systems Engineering, University of Tennessee Knoxville
        {\tt\small khojandi@utk.edu}}%
}

\begin{document}


\maketitle
\thispagestyle{empty}
\pagestyle{empty}

\begin{abstract}
In this paper we propose a novel observer-based method for anomaly detection in  connected and automated vehicles (CAVs). The proposed method utilizes an augmented extended Kalman filter (AEKF) to smooth sensor readings of a CAV based on a nonlinear car-following motion model with time delay, where the leading vehicle's trajectory is used by the subject vehicle to detect sensor anomalies. We use the classic $\chi^2$ fault detector in conjunction with the proposed AEKF for anomaly detection. To make the proposed model more suitable for real-world applications, we consider a stochastic communication time delay in the car-following model. Our experiments conducted on real-world connected vehicle data indicate that the AEKF with $\chi^2$-detector can achieve a high anomaly detection performance. 

\end{abstract}


\section{INTRODUCTION}
Security of connected and automated vehicles (CAVs) has turned into a more critical topic in the recent decade due to the rapid development of intelligent transportation systems that are founded on the connected and automated vehicle technology. A CAV system integrates both connected vehicle (CV) and automated vehicle (AV) technologies, which play a complementary role in improving the operational efficiency, sustainability, and safety of transportation systems \cite{ersal2020connected,van2019path, masoud2017autonomous, abdolmaleki2019itinerary, masoud2016formulations,abdolmaleki2019vehicle ,van2020optimal}.  

Developing an integrated CAV system not only requires equipping vehicles and the transportation infrastructure with a wide array of sensors, actuators and control devices, but also requires vehicle to vehicle (V2V) and vehicle to infrastructure (V2I) communications. Such communications enable controllers that can regulate traffic using network-level information, thereby enhancing mobility in the transportation system; however, the quality of the control strategies become highly dependent on the quality of the input data. As such anomalies in sensor readings, delays in the communication network, or cyber attacks can make such controllers prone to failure, potentially causing unsafe conditions for road users. In addition to informing control strategies, CV data can be used by automated vehicles for trajectory planning. CAVs may fuse data obtained from their onboard sensors as well as data receives through V2V and V2I communications to plan their trajectories. As such, anomalies in the data or cyber attacks targeting their sensor systems can lead to fatal consequences.




A robust anomaly detection scheme designed for CAVs should satisfy three characteristics.
First, any anomaly detection scheme should be able to effectively identify false negatives and false positives, both of which may result in severe negative outcomes as discussed in \cite{wang2019realtime}. To avoid false positives, the anomaly detection methods should be able to distinguish between anomalies and true unexpected changes in network conditions that may trigger unexpected responses from road users. Therefore, to reduce the number of false positives the anomaly detection scheme should be able to incorporate network-level information.  To avoid false negatives, anomaly detection methods should be able to identify the noise, introduced by the vehicles' onboard sensor systems or the communication channel, to make sure decisions are not affected by it. Secondly, anomaly detection methods should not pose additional computational burden on a CAV's already constrained computational and energy resource. Finally, the anomaly detection techniques should be fast enough to be implementable in highly-dynamic traffic streams.

Anomalous sensor readings could be caused by different form of anomalies. Based on previous studies \cite{van2019real, wang2019realtime}, we use the following anomaly taxonomy: ($i$) Short anomaly, which is a short-lived and sudden change in the observed data; ($ii$) Noise, which is a longer-term change (multiple successive readings) in variance of the observed data; ($iii$) Bias, which is an offset from the true sensor readings; ($iv$) Gradual drift, which is a gradual drift in the observed data and ($v$) Miss, which refers to missing data observations that could result from Denial of Service (DoS) attacks or sensor failures. In this paper, we do not explicitly account for the anomaly type `miss'; however, in practice, depending on its duration, the `miss' anomaly type can be viewed as either `instant' or `bias' anomalies, where for a short or long period of time, respectively, the sensor readings are changed to zero. 

The objective of this paper is to develop an anomaly detection scheme that can incorporate network level information, does not require large computational resources, and can detect anomalies in real-time. In our previous work \cite{wang2019realtime}, we developed a framework that combined an adaptive extended Kalman Filter, enhanced using a car-following motion model, and a data-driven fault detector to detect CAV anomalies in the presence of communication time delay. This paper is an extension of our previous work, where we reformulate the problem using an augmented state in order to capture the bias caused by the time delay, and thereby improve the detection performance of the traditional $\chi^2$-detector. 
Moreover, the new method proposed in this paper can be seamlessly coupled with the CNN-KF framework proposed in \cite{van2019real}; experiments indicate  improvements in performance by implementing the new method.
We use the same assumptions as in \cite{wang2019realtime} throughout this paper: ($i$) vehicles follow their immediate upstream vehicle (referred to as the leader), according to a car-following model; and ($ii$) there is a time-delay associated with obtaining the leader's information (e.g., location and velocity).

The rest of the paper is organized as follows: In Section II we provide a brief review of the existing literature in CAV anomaly detection. In Section III  we introduce the problem formulation and our solution method. In section IV we conduct a case study based on a well-known car-following model and compare the performance with our previous method. Finally, in section V we conclude the paper.

\section{Related Work}


Sensor failures and cyber attacks make two of the main contributors to anomalous sensor readings in CAVs. Sensor failure may result from environmental conditions (e.g. dense vegetation and tall buildings that block GPS satellite signals), sensor age, low battery supply, etc. \cite{ni2009sensor}.  
There has been a number of studies focusing on cyber attacks on CAVs \cite{petit2015potential,trippel2017walnut,faughnan2013risk}.
Experimental studies have demonstrated the vulnerability of a wide array of CAV sensors, e.g., speed, acceleration and location sensors, to cyber attacks or faults. For example, a false injection attack  through the CAN bus or the on-board diagnostics (OBD) system of a CAV can result in any of the five anomaly types introduced in this paper \cite{petit2015potential}. As another example, Trippel et al. demonstrate how the in-vehicle acceleration sensor could be vulnerable to acoustic injection attacks \cite{trippel2017walnut}. Spoofing/jamming attacks on a CAV's GPS unit is another example of attackers inducing anomalies into sensor values \cite{faughnan2013risk}. 

Because of the potentially severe consequences of failing to detect anomalous sensor readings and/or anomalous data received through commutation channels, an increasing number of studies have focused on cyber security in CAVs. A wide range of frameworks founded on graph theory \cite{park2015sensor}, deep learning \cite{van2019real}, and game theory \cite{brahmi2019cyber}, among others, have been proposed for this purpose. In this paper we propose a new anomaly detection framework based on extended Kalman filter (EKF) with a car-following motion model and an augmented state space. This framework offers an anomaly detection method that ($i$) directly targets reducing the number of false positive errors in anomaly detection; ($ii$) does not require significant computational resources; and ($iii$) can be executed in real-time, making it suitable for dynamic traffic environments.
This framework is an extension of our previous work \cite{wang2019realtime}, where we focused on detection of anomalous sensor readings and recovery of the corrupt signals. 
Similar to \cite{wang2019realtime}, we use a car-following model as the motion model of an EKF in order to use the trajectory of the subject vehicle's leading vehicle, thereby capturing some network-level information. The leading vehicle's information is received through basic safety messages (BSM) enabled by V2V communications. When using the leading vehicle's information to inform the motion of the subject vehicle, the time-delay in the communication channel could create theoretical challenges that would exclude the use of traditional fault-detectors, such as the $\chi^2$ detector. In our previous work, we addressed this issue by introducing a data-driven fault detector. In this work We adopt an augmented state for EKF, which makes it possible to use a traditional $\chi^2$ detector. 
Additionally, in this paper we analyze the impact of stochastic time delay in receiving the leading vehicle's information. Our experiments show a boost of detection performance of $\chi^2$-detector with augmented state formulation, i.e. augmented extended Kalman filter (AEKF). Additionally, they show a lower mean squared error (MSE) compared with our previous formulation under time delay.

\section{METHODS}

In this section, we first discuss how to reformulate a car-following model into a continuous motion model with time delay using an augmented state. Augmenting the state allows us to compensate for the potential bias caused by the approximation process and model inaccuracy. Based on that, we then formulate a new continuous nonlinear state-space model with discrete measurements based on a car-following model, where the continuous state-transition model represents the intrinsic nature of a vehicle's response to the actions of its immediate leader, and the discrete measurement model represents the discrete nature of sensor sampling. Next, we analyze the impact of stochastic time delay on the system. Finally, we apply the augmented extended Kalman filter (AEKF) to the state-space model, and use the resulting AEKF model in conjunction with a $\chi^2$-detector to find anomalies.

\subsection{Car-Following Model with Time Delay}
We use a typical car-following model as described in \cite{treiber2014traffic}:

\begin{equation}\label{eq:1}
\begin{aligned}
d_n(t) &= x_{n-1}(t)-x_n(t)\\
\dot{d}_n(t) &= v_{n-1}(t) - v_n(t)\\
\dot{v}_{n}(t) &= f(v_{n}(t-\tau),{d_n(t-\tau)}, {\dot{d}_n(t-\tau)})\\
\end{aligned}
\end{equation}
where $\dot{v}_{n}(t)$, $v_n(t)$, $x_n(t)$ are the acceleration, speed, and location of the $n$th vehicle, to which we refer as the subject vehicle or the following vehicle, respectively. In Eq. \ref{eq:1}, $d_n(t)$ and $\dot{d}_n(t)$ are the headway and the speed difference between the subject vehicle and its leading vehicle (also referred to as the leader), i.e. the $(n-1)$th vehicle, respectively. Parameter $\tau$ denotes time delay, intended to capture the time lapsed between the moment the leader performs an action, to the moment the subject vehicle acts in response. For now we consider a constant time delay $\tau$; however, later we relax the constant assumption by considering a stochastic time delay.
Function $f$ is the stimulus function. 

Since we are analyzing the subject vehicle's sensors, we define the state vector of the $n$th vehicle in continuous time as $s_n(t) = \left[x_{n}(t),v_{n}(t)\right]^T \in \mathbb{R}^{2}$.
%
%
Similarly, we define the input vector containing information received from the leading vehicle as $u_{n}(t) = [x_{n-1}(t),v_{n-1}(t)]^T$.
%
Consequently, we can recast Eq. (\ref{eq:1}) as a function of $s_n(t)$, producing a car-following model that maps the state into an actionable decision for the subject vehicle:
\begin{equation}\label{eq:2.1}
\begin{aligned}
\dot{v}_{n}(t) 
:= &f_{v}\left(s_n(t-\tau),u_{n}(t-\tau))\right.
\end{aligned}
\end{equation}
where $f_{v}$ denotes the stimulus function describing velocity in a continuous sate space.

\subsection{State-Space Model with Continuous State and Discrete Measurement}
In this section we define a state-space model with a continuous state-transition model and discrete measurements, which will be used for AEKF. Based on previous definition of the state vector $s_n(t)$, the state-transition model satisfies the following differential equation:

\begin{equation}\label{eq:11}
\dot{s}_n(t) = 
\begin{bmatrix} 
\dot{x}_n(t)  \\
\dot{v}_n(t)  
\end{bmatrix} 
= 
\begin{bmatrix}
e_2^T s_n(t)\\
f_{v}(s_n(t-\tau),u_{n}(t-\tau))
\end{bmatrix},
\end{equation}
where $e_2 = [0,1]^T$ is the standard basis vector.

Ideally when $\tau = 0$, the state-space model in equation (\ref{eq:11}) satisfies the Markovian property, allowing for applying AEKF. However, in practice, $\tau$ is usually a nonzero value affected by various factors, e.g., communication network or data processing delay, etc.. As such, in practice AEKF cannot be applied to equation (\ref{eq:11}), since the derivative of the state vector is determined by multiple previous state vectors. 

In order to apply AEKF, we approximate Eq. (\ref{eq:11}) in the following way: 
Based on the bounded acceleration assumption, we can obtain a delay differential equation (DDE), describing the delayed state-transition model:
\begin{equation}\label{eq:12}
\begin{aligned}
\dot{s}_n(t) &= 
\begin{bmatrix} 
\dot{x}_n(t)  \\
\dot{v}_n(t)  
\end{bmatrix} \\
&= 
\begin{bmatrix}
e_2^T s_n(t-\tau) + \int_{t-\tau}^{t}a_n(r)dr\\
f_{v}(s_n(t-\tau),u_{n}(t-\tau))
\end{bmatrix}\\
&= 
\begin{bmatrix}
1,0,1\\
0,1,0\\
\end{bmatrix}
\times
\begin{bmatrix}
\Tilde{e}_2^T s_n(t-\tau)\\
f_{v}(s_n(t-\tau),u_{n}(t-\tau))\\
\int_{t-\tau}^{t}a_n(r)dr
\end{bmatrix} \\
&=
\begin{bmatrix}
1,0,1\\
0,1,0\\
\end{bmatrix}
\times 
\begin{bmatrix} 
\Tilde{e}^T_2 \Tilde{s}_n(t-\tau)  \\
f_{v}(\Tilde{e}_{12}^T\Tilde{s}_n(t-\tau),u_n(t-\tau))  \\
\Tilde{e}_3^T\Tilde{s}_n(t-\tau)
\end{bmatrix} 
\end{aligned}
\end{equation}
where $a_n(t)$ is the acceleration of the $n$th vehicle at time $t$, and 
$\Tilde{s}_n(t) =[{x}_n(t),{v}_n(t),{\delta}_n(t)]^T$ denotes the augmented state vector of $s_n(t)$ with augmented state $\delta_n(t) = \int_{t-\tau}^t a_n(r)dr$, $\Tilde{e}_{12} = [\Tilde{e}_1, \Tilde{e}_2, 0]^T$, and $\Tilde{e}_i \in \mathbb{R}^3$ is the standard basis vector with $i$th element equal to 1 and 0 otherwise. Since $\delta_n(t)$ is unknown, we assume it is a constant or a random variable with small variance. Thus we have $\dot{\delta}_n(t) \approx 0$.

The state-transition model with respect to the augmented state vector $\Tilde{s}_n(t)$ can be presented as follows:

\begin{equation}
    \label{eq:state_transit}
    \begin{aligned}
    \dot{\Tilde{s}}_n(t) &= 
    \begin{bmatrix}
    \dot{x}_n(t)\\
    \dot{v}_n(t)\\
    \dot{\delta}_n(t)
    \end{bmatrix}\\ 
    &= \begin{bmatrix}
    \Tilde{e}^T_2 \Tilde{s}_n(t-\tau) + \Tilde{e}_3^T\Tilde{s}_n(t-\tau)\\
    f_{v}(\Tilde{e}_{12}^T\Tilde{s}_n(t-\tau),u_n(t-\tau))\\
    0
    \end{bmatrix} 
    + \theta(t)\\
    &= g(\Tilde{s}_n(t-\tau),u_{n}(t-\tau))+{\theta(t)}
    \end{aligned}
\end{equation}
where $g(\cdot)$ is the motion model and ${\theta(t)}$ is the process noise, which accounts for the error introduced by the approximation and model inaccuracy.

Finally, using the new augmented state vector $\Tilde{s}_n(t)$, we obtain a continuous-time state-transition model with discrete-time measurement as the following:
\begin{equation}\label{eq:13}
\begin{aligned}
\dot{\Tilde{s}}_n(t) &= g(\Tilde{s}_n(t-\tau),u_{n}(t-\tau))+{\theta(t)}\\
z_n(t_k) &= h(\Tilde{s}_n(t_k))+{\eta(t_k)},\ k \in \{0 \cup \mathbb{Z}^+\}\\
\end{aligned}
\end{equation}
where $h(\cdot)$ is the measurement function, $z_n(\cdot)$ denotes sensor reading of the leading vehicle, ${\eta(t_k)}$ is the observation noise, which is assumed to be mutually independent with the process noise, $t_{k+1} = t_k + \Delta t, ~k \in \{0 \cup \mathbb{Z}^+\}$, and $\Delta t$ is the sampling time interval for sensors.

\subsection{Stochastic Time Delay}
Now we consider a more general case where the time delay $\tau_s$ is not a known constant. We assume that $\tau_s$ is a linear model:
\begin{equation}\label{eq:random delay}
    \tau_s = \tau + \kappa,
\end{equation}
where $\kappa$ is zero mean Gaussian distributed with variance $\sigma_1^2$. Therefore we have $E(\tau_s) = \tau$. We also assume the leading vehicle's 
trajectory obeys a linear model, i.e.,
\begin{equation}
    \begin{cases}
    \dot{x}_{n-1}(t) = v_{n-1}(t)\\
    \dot{v}_{n-1}(t) = a_{n-1}(t).
    \end{cases}
\end{equation}
Since the leading vehicle's motion model is unknown, we assume that the acceleration of the leading vehicle, $a_{n-1}$, is a Gaussian process with zero mean and variance $\sigma_2^2$.

 \emph{Claim:} A random time delay $\tau_s$ is equivalent to adding noise into the input vector $u_{n}(t-\tau)$ with fixed time delay $\tau$.

\begin{proof}
By integrating $\dot{x}_{n-1}(t-\tau_s)$, we have
\begin{equation}
    \begin{aligned}
    x_{n-1}(t-\tau_s) &= \int_{0}^{t-\tau_s} v_{n-1}(\xi) d\xi\\
    &= \int_{0}^{t-{\tau}-\kappa} v_{n-1}(\xi) d\xi\\
    &=\int_{0}^{t-\tau} v_{n-1}(\xi) d\xi - \int_{t-\tau - \kappa}^{t-\tau} v_{n-1}(\xi) d\xi \\
    &= x_{n-1}(t-\tau) + \epsilon_1(\kappa,t-\tau)
    \end{aligned}
\end{equation}
where $\epsilon_1(\theta,t-\tau) = -\int_{t-\tau - \theta}^{t-\tau} v_{n-1}(\xi) d\xi$. Therefore using random time delay is equivalent to adding noises $\epsilon_1(\kappa,t-\tau)$ into state $x_{n-1}(t-\tau)$. Note that $\epsilon_1(\kappa,t-\tau)$ does not necessarily have zero mean:
\begin{equation}
  \textrm{E}_{\kappa}[\epsilon_1(\kappa,t-\tau)] = \int_{-\infty}^{\infty}\phi(\kappa)\int_0^{-\kappa}v_{n-1}(\xi+t-\tau)d\xi d\kappa  
\end{equation}
where $\phi$ is the Gaussian probability density function.

Similarly, for $v_{n-1}$, using random time delay is equivalent to adding noise $\epsilon_2(\kappa,t-\tau)$ into $v_{n-1}(t-\tau)$:

\begin{equation}
    v_{n-1}(t-\tau_s) = v_{n-1}(t-\tau) + \epsilon_2(\kappa,t-\tau)
\end{equation}
where $\epsilon_2(\kappa,t-\tau)$ does not necessarily have zero mean.

Therefore, the input vector can be expressed as a linear model:
\begin{equation}\label{eq:add noise}
    u_{n}(t-\tau_s) = u_{n}(t-\tau) + \epsilon(\kappa,t-\tau),
\end{equation}
where $\epsilon(\kappa,t-\tau) = [\epsilon_1(\kappa,t-\tau), \epsilon_2(\kappa,t-\tau)]^T$. 

\end{proof}


When the time delay of the input vector is stochastic, the continuous-time  state-transition model with discrete-time measurement is as follows:
\begin{equation}\label{eq:15}
\begin{aligned}
\dot{\tilde{s}}_n(t) &= g(\tilde{s}_n(t-\tau),u_{n}(t-\tau_s))+\theta(t)\\
z_n(t_k) &= h(\tilde{s}_n(t_k))+\eta(t_k),\ k \in \{0 \cup \mathbb{Z}^+\}.\\
\end{aligned}
\end{equation}
Because of the term $\epsilon(\theta,t-\tau)$, plugging equation (\ref{eq:add noise}) into (\ref{eq:15}) could cause a non-zero mean for $\theta(t)$, depending on the specific formulation of the car-following model. As mentioned in the next section, the existence of bias in the process noise $\theta(t)$ could deteriorate the performance of the traditional $\chi^2$ fault detector, whereas using AEKF can mitigate this issue.

\subsection{Augmented Extended Kalman Filter with Fault Detector}

In order to smooth the CAV sensor noise, we apply AEKF to the state-space model in Eq. (\ref{eq:13}). Note that for the ease of notation, we use state vector notation $s$ instead of the augmented state vector $\tilde{s}$, and we also omit subscript $n$ for simplicity. AEKF consists of the following 3 steps:

\emph{Step 0 - Initialize state mean and covariance:}
\begin{equation} \label{eq:7}
\begin{aligned}
    &\hat{s}_{k|k-1} = \textrm{E}[s(t_0)]\\
    & P_{k|k-1} = \textrm{Var}[s(t_0)].\\
\end{aligned}
\end{equation}

\emph{Step 1 - Predict state and state covariance:}
\begin{equation} \label{eq:7.1}
    \begin{aligned}
    \textrm{Solve}
    &\begin{cases}
        \dot{\hat{s}}(t) = g(\hat{s}(t-\tau),u(t-\tau)),\\
        \dot{P}(t) = F(t-\tau)P(t-\tau)+P(t-\tau)F(t-\tau)^T + Q(t)\\
    \end{cases}\\
    \textrm{with}
    &\begin{cases}
        \hat{s}(t_{k-1}) = \hat{s}_{k-1|k-1}\\
        P(t_{k-1}) = P_{k-1|k-1}
    \end{cases}\\
    \Rightarrow 
    &\begin{cases}
        \hat{s}_{k|k-1} = \hat{s}(t_k)\\
        P_{k|k-1} = P(t_k)
    \end{cases}
    \end{aligned}
\end{equation}
where $F(t-\tau) = \frac{\partial g}{\partial s}|_{\hat{s}(t-\tau),u(t-\tau)}$ is the first-order approximation of the Jacobian matrix of function $g(\cdot)$.

\emph{Step 2 - Update state and state covariance:}
\begin{equation} \label{eq:8}
\begin{aligned}
    &\nu_k = z(t_k) - h(\hat{s}_{k|k-1})\\
    &S_k = H(t_k) P_{k|k-1}H(t_k)^T + R_k\\
    &K_k = P_{k|k-1}H(t_k)^T S_k^{-1}\\
    &\hat{s}_{k|k} =\hat{s}_{k|k-1}+ K_k\nu_k\\
    &P_{k|k} = P_{k|k-1} - K_k H(t_k)P_{k|k-1}\\
\end{aligned}
\end{equation}
where 
$H(t_k) = \frac{\partial h}{\partial s}|_{\hat{s}_{k|k-1}}$, $Q(t)$ is the covariance matrix of the process noise at time $t$, $R_k = R(t_k)$ is the covariance matrix of the measurement noise at time $t_k$, and $\nu_k$ is innovation, which is the difference between the measurement and the prediction, at time $t_k$. 

One of the classic fault detectors used in conjunction with Kalman filter is the $\chi^2$-detector \cite{bar1995multitarget}. Since AEKF is essentially a special type of Kalman filter, the $\chi^2$-detector can be seamlessly applied to AEKF as well. Specifically, it constructs $\chi^2$ test statistics to determine whether the new measurement falls into the gate region with the probability determined by the gate threshold in Eq. \ref{eq:chi2d},
\begin{equation}\label{eq:chi2d}
\begin{aligned}
V_{\gamma}(k) = \{z:&(z - \hat{z}_{k|k-1})^T S_k^{-1} (z-\hat{z}_{k|k-1})\leq \sigma\},\\
\end{aligned}
\end{equation}
where $\hat{z}_{k|k-1}$ is the predicted value of measurement at time $t_k$.
The $\chi^2$ test statistics for the fault detector is defined as $\chi^2(t_k)= \nu_k^T S_k^{-1}\nu_k$.
 In order to make the $\chi^2$ test provide meaningful results, the innovation $\nu_k$ should be zero mean Gaussian distributed with covariance $S_k$; therefore, we combine it with AEKF, which could compensate the potential bias caused by time delay.

In summary, we combine AEKF and $\chi^2$ fault detector to detect anomalies and recover the corrupt sensor readings. Specifically, at each time epoch, the subject vehicle receives the measurements from both the leading vehicle and its own onboard sensors. AEKF uses the car-following motion model to smooth the following vehicle's speed and location signals. During the smoothing process, the AEKF generates the innovation, which is the the discrepancy between the measurements and the prediction, and sends the innovation to the fault detector model for anomaly detection. If there is no sensor anomaly detected, the innovation will be combined with the measurement at the current time epoch in order to generate an estimation. Otherwise, the prediction would replace the estimation, which will be used in the next time epoch.


\section{Case Study 
}
In this section we use the well-known Intelligent Driver Model (IDM) proposed by Treiber et al. \cite{treiber2000congested} as our car-following model of choice, and implement our framework to compare the anomaly detection performance of the $\chi^2$-detector in conjunction with EKF and AEKF. Since the IDM has no explicit reaction time and its driving behavior is given in terms of a continuously differentiable acceleration function \cite{treiber2014traffic}, it is suitable for modeling semi-automated (compared to human) driving. However, it can be easily extended to capture the communication delay, as described in the previous section. Note that in order to evaluate system performance, we assume that the input vector containing the leading vehicle's information is not anomalous. We also use  a Gaussian distribution with mean $\tau$ and variance 1 to generate stochastic time delay.

Using the definition of the augmented state $\tilde{s}_n(t)$ and the input $u_n(t)$ in the previous section, the IDM model with fix time delay $\tau$ can be described as the following:
\begin{equation*}
 \label{eq:21}
\begin{aligned}
\dot{x}_n(t) &= v_n(t) \\
\dot{v}_n(t) &= f_{v}(\Tilde{e}_{12}^T \tilde{s}_n(t-\tau),u_n(t-\tau)) \\
&=a\left(1-\left(\frac{v_n(t-\tau)}{v_0}\right)^{\delta}\right.\\
&-\left.\left(\frac{s^*(v_n(t-\tau),v_n(t-\tau)-v_{n-1}(t-\tau))}{x_{n-1}(t-\tau)-x_n(t-\tau)-l_n}\right)^2\right)
\end{aligned}
\end{equation*}
with
$$
s^*(v_n,\Delta v_n) = s_0 + v_nT+\frac{v_n\Delta v_n}{2\sqrt{ab}} 
$$
where $a, b, \delta, v_0, s_0, T$ and $l_n$ are model parameters. The state vector $\tilde{s}_n$ has dimension of 3 and the input vector $u_n$ has dimension of 2. For detailed information on IDM refer to \cite{treiber2000congested}. Following the typical parameter values of city traffic used in \cite{treiber2014traffic}, we set the parameter values in our study as follows: $a = 1.0, b = 1.5, \delta = 4, v_0 = 33.75, s_0 = 2, T = 1.0$, $l_n = 5$, and define the measurement function $h(\cdot)$ of AEKF in equation (\ref{eq:15}) as:
\[h(\tilde{s}) = H \cdot \tilde{s} = 
\begin{bmatrix}
1,0,1\\
0,1,0\\
\end{bmatrix} \cdot \tilde{s}.\]

\begin{figure*}[t!]
    \centering
    \begin{subfigure}[t]{0.35\textwidth}
        \centering
        \includegraphics[width = 0.9\linewidth]{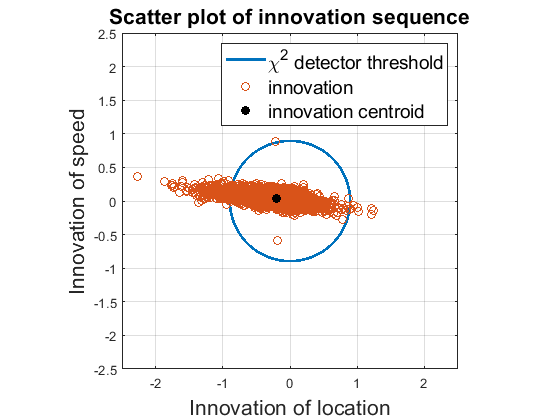}
        \caption{Scatter plot of EKF.}
        \label{fig2.1}
    \end{subfigure}%
    \begin{subfigure}[t]{0.35\textwidth}
        \centering
        \includegraphics[width = 0.9\linewidth]{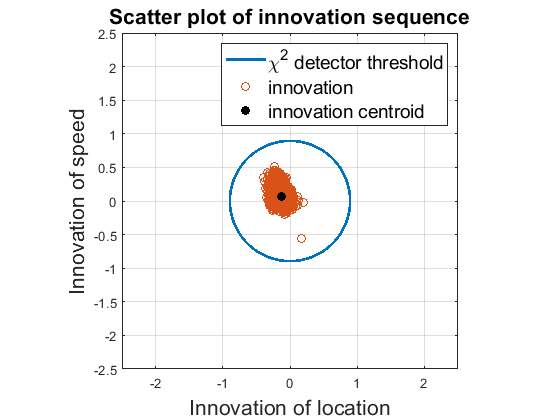}
        \caption{Scatter plot of AEKF.}
        \label{fig2.2}
    \end{subfigure}
    \caption{Example of a normalized innovation sequence being non-zero mean, where $\tau = 1.5$ seconds. The threshold $\sigma$ of $\chi ^2$-detector is 0.8.}
    \label{fig2}
\end{figure*}

We use the same dataset as in \cite{wang2019realtime}, which is based on Safety Pilot Model Deployment (SPMD) program \cite{bezzina2014safety} funded by the US department of Transportation, and collected in Michigan. The anomalies are randomly generated with 5\% anomaly rate and injected into the raw dataset using algorithm 1 in \cite{wang2019realtime}.

We implement two sets of models, where model 1 is composed of the traditional EKF with an IDM motion model and a $\chi^2$-detector, and model 2 is composed of the AEKF with the IDM model and the $\chi^2$-detector. Each model is implemented under three experimental settings generated by varying the value of the anomaly parameter $c_i, i=\{1,2,3,4\}$ of algorithm 1 in \cite{wang2019realtime}, which depending on the anomaly type describes the variance or magnitude of different types of anomalies. More specifically, values of $c_i=1$, $c_i=0.5$, and $c_i=0.1$ are used for settings 1, 2 and 3, respectively. This suggests anomalous readings become more subtle, and generally more difficult to detect, from setting 1 to setting 3. We also conduct the each experimental setting under 3 different time delays of $\tau = \{0,0.5,1.5\}$ (seconds). Lastly, the maximum duration of an anomaly, $L$, is set to 20 for each setting. In order to evaluate the impact of changing models/parameters, we compute the Area Under the Curve (AUC) for each receiver operating characteristic (ROC) curve. 


\begin{table}[h]
\centering
\caption{AUC of detection performance and its standard deviation across 20 different executions for two models, at the anomaly rate of  5\% and in the presence of all anomaly types. P-values indicate statistical significance at 5\% level using paired $t$-test between the detection performance of each pair of models except between for the scenarios with $c_i =1$ and $\tau=0.5$. }
\label{tab:1}
\begin{tabular}{c|ccc|ccc}
\hline
                                                           & \multicolumn{3}{c|}{\textbf{Model 1}}                                                           & \multicolumn{3}{c}{\textbf{Model 2}}                                                          \\ \hline
\textbf{$\tau$}                                            & \textbf{0.0}                   & \textbf{0.5}                   & \textbf{1.5}                    & \textbf{0.0}                   & \textbf{0.5}                   & \textbf{1.5}                   \\ \hline
\multirow{2}{*}{\textbf{$c_i = 1$}}                        & 0.984                          & 0.931                          & 0.855                           & 0.983                          & 0.945                          & 0.902                          \\
                                                           & $\pm$0.015                     & $\pm$0.028                     & $\pm$0.034                      & $\pm$0.022                     & $\pm$0.030                     & $\pm$0.038                     \\ \hline
\multirow{2}{*}{\textbf{$c_i = 0.5$}}                      & 0.972                          & 0.917                          & 0.799                           & 0.969                          & 0.928                          & 0.844                          \\
                                                           & \multicolumn{1}{l}{$\pm$0.020} & \multicolumn{1}{l}{$\pm$0.033} & \multicolumn{1}{l|}{$\pm$0.391} & \multicolumn{1}{l}{$\pm$0.024} & \multicolumn{1}{l}{$\pm$0.034} & \multicolumn{1}{l}{$\pm$0.046} \\ \hline
\multicolumn{1}{l|}{\multirow{2}{*}{\textbf{$c_i = 0.1$}}} & \multicolumn{1}{l}{0.871}      & \multicolumn{1}{l}{0.718}      & \multicolumn{1}{l|}{0.576}      & \multicolumn{1}{l}{0.867}      & \multicolumn{1}{l}{0.759}      & \multicolumn{1}{l}{0.731}      \\
\multicolumn{1}{l|}{}                                      & \multicolumn{1}{l}{$\pm$0.029} & \multicolumn{1}{l}{$\pm$0.031} & \multicolumn{1}{l|}{$\pm$0.034} & \multicolumn{1}{l}{$\pm$0.035} & \multicolumn{1}{l}{$\pm$0.039} & \multicolumn{1}{l}{$\pm$0.058} \\ \hline
\end{tabular}
\end{table}

Table \ref{tab:1} presents the AUC values, averaged over 20 random instances, of all nine scenarios that are generated by changing the intensity of anomalies ($c_i= \{1,0.5,0.1\}$) and time delay ($\tau=\{0,0.5,1,5\})$. For each scenario, we use a paired $t$-tests with a 5\% significant level to determine whether there is a statically significant difference in the AUC values of Models 1 and 2.
Results indicate that when there is no time delay, EKF with $\chi^2$-detector (model 1) consistently achieves a better anomaly detection performance than the AEKF with $\chi^2$-detector (model 2). This is because when time delay is zero, there is no need to compensate for potential bias. Under such circumstances, the augmented state variable behaves similarly to additional noise added to the first element of the state variable (i.e., geo-location) as shown in Eq. (\ref{eq:state_transit}), causing a lower detection performance.
Results indicate that model 2 outperforms model 1 in all scenarios where $\tau>0$, by capturing the the bias that is generated due to time-delay. 
It should be noted that in practice, a zero time delay may rarely occur, indicating that the augmented state formulation would perform better in practice.
Results also indicate that there is a degeneracy of performance for each method as the parameter $c_i$ becomes smaller. This observation is in line with intuition, since smaller $c_i$ makes the anomaly more subtle and therefore harder to detect. Additionally, the trends of AUC values indicate that as we increase the time delay, the overall detection performance systemically deteriorates. This suggests that the time delay in general may have a negative impact on detection performance.
 
 \begin{table}[h]
 \centering
\caption{Mean innovation for each state variable and the MSE for the two models at zero anomaly rate. MSE is calculated as the squared root of sum of the MSEs for the two state variables. }
\label{tab:2}
\begin{tabular}{cccc}
\hline
\multicolumn{1}{c|}{\textbf{$\tau$}}        & \textbf{0.0} & \textbf{0.5} & \textbf{1.5} \\ \hline
\multicolumn{4}{c}{\textbf{Model 1}}                                                     \\ \hline
\multicolumn{1}{c|}{\textbf{Mean innovation value of location}} & -0.003       & -0.146       & -0.436       \\ \hline
\multicolumn{1}{c|}{\textbf{Mean innovation value of speed}}    & 0.022        & 0.028        & 0.062        \\ \hline
\multicolumn{1}{c|}{\textbf{MSE}}           & 7.12E-04     & 0.033        & 0.098        \\ \hline
\multicolumn{4}{c}{\textbf{Model 2}}                                                     \\ \hline
\multicolumn{1}{c|}{\textbf{Mean innovation value of location}} & -0.002       & -0.033       & -0.097       \\ \hline
\multicolumn{1}{c|}{\textbf{Mean innovation value of Speed}}    & 0.024        & 0.036        & 0.095        \\ \hline
\multicolumn{1}{c|}{\textbf{MSE}}           & 8.43E-04     & 0.032        & 0.095        \\ \hline
\end{tabular}
\end{table}

In order to investigate the effect of using the augmented state formulation, we find the mean innovation value for the two state variables (i.e., location and speed) for each model under the three time delay settings with zero percentage of anomaly, as presented in TABLE \ref{tab:2}. As shown in the TABLE \ref{tab:2}, AEKF can significantly decrease the background bias in the first state variable, i.e. location of the subject vehicle, when there is a time delay. Similar to previous results, when the time delay is zero, the performance of AEKF deteriorates, since no potential bias need to be compensated for, and the augmented state variable only introduces more randomness in the system. As time delay becomes larger, the effect of bias correction becomes more prominent. TABLE \ref{tab:2} also shows the mean squared error (MSE) in the innovation of the two models, where MSE is calculated as the squared root of sum of the MSEs for the two state variables. further confirming our previous observations--as we increase the time delay, the reduction in MSE under model 2 (AEKF) becomes larger compared with that under model 1 (EKF). Figures \ref{fig2.1} and \ref{fig2.2} display the scatter plots of normalized innovation for both EKF and AEKF, generated from the training dataset in our experiments, when there exits a time delay and $\int_{t-\tau}^{t}a_n(r)dr$ in Eq. (\ref{eq:12}) is not zero mean. In the 2-dimensional case, the $\chi^2$-detector defines a circular boundary with its center located at $(0,0)$, i.e., the blue lines in figure \ref{fig2}, which corresponds to the thresholding boundary of the $\chi^2$-detector with $\sigma =0.8$. We can see a significant reduction in the mean and variance of the normalized innovation of the first state variable in AEKF. This figure also shows that the innovation sequence of AEKF resembles a normal distribution more closely, indicating the suitability of the $\chi^2$-detector.

\section{CONCLUSIONS}

The goal of this study is to develop a framework to detect anomalous sensor readings and/or data received through V2V or V2I communications in CAVs. The proposed framework introduces a computationally light-weight anomaly detection tool that attempts to minimize false positives and negatives by incorporating context from the driving environment, obtained through V2V communications. The delay in the communication channel can introduce theoretical challenges that preclude using existing anomaly detection tools. As such, we introduce an augmented extended Kalman filer (AEKF) that is informed by not only the vehicle's onboard sensors but also the leading vehicle's trajectory. In our case study, we use the well-known intelligent driver car following model to incorporate the leading vehicle's information. In conjunction with AEKF, we use the classic $\chi^2$-detector to detect five types of anomalies, which encapsulate general sensor faults and/or cyber attacks in CAVs. We also analyze the effect of stochastic time delay on the detection performance. We quantify the effect of these contributions by conducting experiments under two scenarios: $\chi^2$-detector with an IDM motion model and EKF, and $\chi^2$-detector with an IDM  motion model and AEKF. Results indicate that in the presence of time delay, the second model outperforms the first. Furthermore, results show that in the presence of time delay, AEKF can decrease the overall innovation MSE by compensating for the background bias. 
\vspace{0.2cm}

\noindent\textbf{ACKNOWLEDGMENT}: This study was partially supported by grant \#69A3551747105 from the US DOT center for connected and automated transportation.

\bibliographystyle{IEEEtran}
\bibliography{IEEEabrv,Refs}

\addtolength{\textheight}{-12cm}   










\end{document}